\documentclass[letterpaper]{article} 
\usepackage{aaai2027}  
\usepackage[hyphens]{url}  
\usepackage{graphicx} 
\usepackage{natbib}  
\usepackage{caption} 
\usepackage{algorithm}
\usepackage{algorithmic}
\usepackage[hyphens]{url}
\usepackage{graphicx}
\usepackage{natbib}
\usepackage{caption}
\usepackage{booktabs}
\usepackage{amsmath,amssymb}
\usepackage{multirow}
\usepackage{microtype}
\usepackage{bm}
\usepackage{mathtools}
\usepackage{amsthm}

\newtheorem{proposition}{Proposition}

\newcommand{\MRA}{\mathrm{MRA}}
\newcommand{\MRAret}{\mathrm{MRA}_{\mathrm{ret}}}
\newcommand{\Cov}{\mathrm{Cov}}
\newcommand{\Scps}{s_{\mathrm{cps}}}
\newcommand{\Svp}{s_{\mathrm{vp}}}

\usepackage{xcolor}
\newcommand{\fh}[1]{{\color{purple}{}}}
\usepackage{newfloat}
\usepackage{listings}
\DeclareCaptionStyle{ruled}{labelfont=normalfont,labelsep=colon,strut=off} 
\floatstyle{ruled}
\newfloat{listing}{tb}{lst}{}
\floatname{listing}{Listing}

\title{Answer-Level Trust Selection for Physical Vision-Language Reasoning}
\author{
    Rongyu Yu\textsuperscript{\rm 1},
    Ke Niu\textsuperscript{\rm 2},
    Fengxiang He\textsuperscript{\rm 1}
}
\affiliations{
    \textsuperscript{\rm 1}University of Edinburgh\\
    \textsuperscript{\rm 2}Fudan University\\
    v1ryu33@ed.ac.uk, 22110240103@m.fudan.edu.cn, F.He@ed.ac.uk
}

\begin{document}

\maketitle

\begin{abstract}
Vision-language models (VLMs) can estimate physical quantities such as
duration, speed, and acceleration from visual observations, but existing
benchmarks primarily assess overall model performance against annotated
ground truth. In deployment, a key question is whether an individual
prediction can be trusted when its ground truth is unavailable.
Self-consistency alone may fail to capture important failure modes: a VLM
may produce stable-but-wrong estimates or rely on textual priors rather than
visual evidence. We formulate answer-level selective prediction for quantitative physical
reasoning and propose Answer-Level Trust Selection (ATS), a post-hoc,
model-agnostic framework for accepting or rejecting individual VLM
predictions. ATS requires no fine-tuning, auxiliary verifier, or access to
the model's internal logits. Instead, it aggregates eight interpretable
behavioral diagnostic scores derived from repeated queries and controlled
interventions into a unified trust score. We evaluate ATS in depth on Qwen2.5-VL-7B and across 20 VLM backbones,
examining selective performance, diagnostic behavior, and targeted failure
modes. Our results show that intervention-based diagnostics help identify
stable-but-wrong and prior-tracking predictions that repeated agreement
alone may miss. However, improved failure-case rejection can come at the
cost of lower retention of correct predictions. ATS therefore complements
model-level capability evaluation with answer-level reliability assessment
for quantitative VLM predictions. Code will be released upon publication.
\end{abstract}

\section{Introduction}
\begin{figure}[t]
    \centering
    \includegraphics[
        width=1.0\columnwidth,
        trim=25 15 15 20,
        clip
    ]{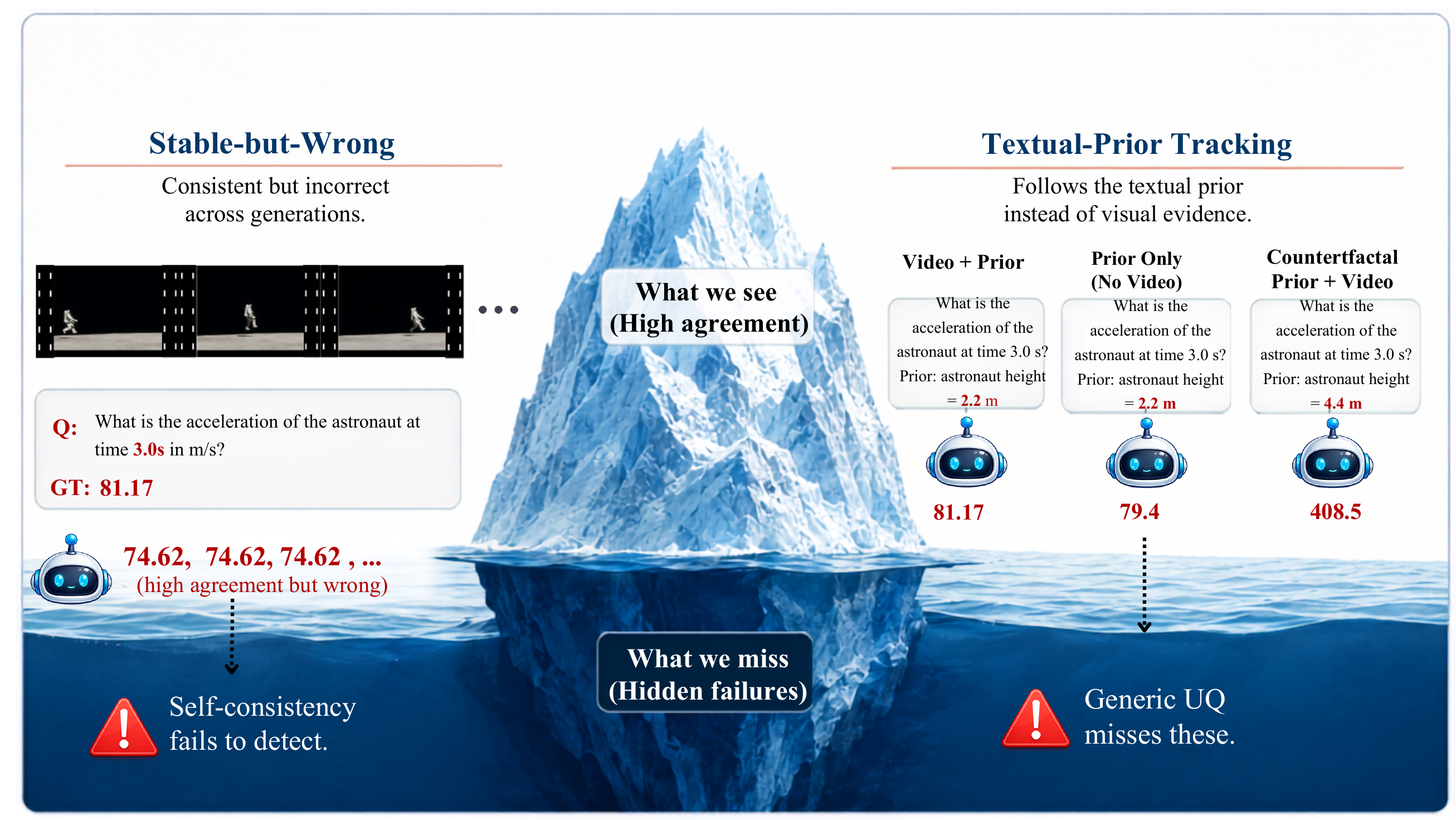}
\caption{
Hidden failure modes in physical VLM reasoning. Repeated agreement can
conceal stable-but-wrong answers, while predictions may also follow perturbed
textual priors rather than remain stable under fixed visual evidence.
}
    \label{fig:1}
\end{figure}

Vision-language models (VLMs) are increasingly used as perception and
reasoning modules in systems that interact with the physical
world~\cite{chow2025physbench,puyin2026quantiphy}. Beyond object recognition
and scene description, these models are now expected to reason about measurable
physical quantities such as size, duration, speed, and
acceleration~\cite{chow2025physbench,puyin2026quantiphy}. Their numerical
predictions may serve as interfaces between visual perception and downstream
planning, monitoring, measurement, and decision-support systems. This shift
makes quantitative physical reasoning an important test of whether VLM outputs
are not only semantically plausible but also sufficiently reliable for
downstream use. Recent benchmarks such as
QuantiPhy~\cite{puyin2026quantiphy} evaluate whether VLMs can infer physical
quantities of moving objects from visual observations, extending VLM
evaluation from scene description to numerically grounded visual reasoning.

However, existing benchmarks for quantitative VLM reasoning primarily evaluate
model-level performance by comparing predictions with annotated ground truth
and averaging accuracy over a dataset. Although such evaluation
is essential for measuring overall capability, deployment raises a more
actionable question: once a VLM produces a particular numerical estimate, is
that specific answer sufficiently trustworthy to retain? As illustrated in
Figure~\ref{fig:1}, numerical VLMs may exhibit hidden
answer-level failures. A model can be stable but wrong, repeatedly producing
similar numerical estimates that remain far from the ground truth. It may also
be highly sensitive to textual numerical priors: when the visual input is held
fixed but the prior is perturbed, its prediction may shift toward the altered
prior. Average benchmark performance cannot identify which individual answers
exhibit these behaviors at deployment time, while agreement across repeated
outputs may also fail to expose them.

These limitations motivate a complementary perspective on evaluating the
reliability of numerical VLM predictions. Building on the principle of
selective prediction, an answer-level selection mechanism should retain a
prediction only when its estimated reliability is sufficient and otherwise
abstain from using it.
Unlike conventional model-level evaluation, which characterizes overall
capability using aggregate benchmark metrics, this answer-level perspective
focuses on whether a particular prediction should be accepted or rejected when
its ground truth is unavailable at deployment time. The two perspectives are
complementary: model-level evaluation measures how well a VLM performs on
average, whereas answer-level selection supports operational decisions about
individual outputs. This distinction directly addresses the practical need to
separate numerical answer generation from post-hoc accept-or-reject decisions.

We propose {Answer-Level Trust Selection (ATS)}, a post-hoc, black-box
selective trust framework for quantitative physical reasoning with VLMs. ATS
offers two practical advantages. First, it is model-agnostic and minimally
invasive: it requires no fine-tuning of the VLM, no additional neural
verifier, no access to internal logits, and no modification or replacement of
the original answer. Second, ATS provides an interpretable basis for
answer-level selection. It evaluates each prediction using eight behavior-based diagnostic scores, each capturing a distinct aspect of the
model's predictive behavior, and aggregates them into a unified trust score for
accepting or rejecting the original answer. We conduct an in-depth analysis of
ATS using Qwen2.5-VL-7B as the primary model, examining its selective
performance and the contribution of each diagnostic component. To further
characterize the current state of quantitative physical reasoning, we
additionally benchmark 20 VLMs, providing a broad comparison of their numerical
reasoning capabilities.

Our main contributions are summarized as follows:
\begin{itemize}
    \item
    
We formulate answer-level selective prediction for quantitative physical
reasoning with VLMs, focusing on whether to accept or reject individual
answers in real-world deployment.

\item We propose {Answer-Level Trust Selection (ATS)}, a post-hoc,
black-box framework. ATS is model-agnostic and minimally
invasive, while providing an interpretable basis for answer selection.

    \item We evaluate ATS in depth on Qwen2.5-VL-7B and analyze its
    performance. We further
    benchmark 20 representative VLMs to characterize their capabilities in quantitative physical reasoning.
\end{itemize}

\subsection{Quantitative Physical Reasoning in VLMs}

Visual physical reasoning benchmarks evaluate models' understanding of object
interactions, dynamics, causal relations, and latent physical properties.
CLEVRER~\cite{yi2020clevrer} and Physion~\cite{bear2021physion} study
collision reasoning and physical scene evolution, while
ComPhy~\cite{chen2022comphy} and CRIPP-VQA~\cite{patel2022cripp} require
inference of properties such as mass, charge, friction, and initial velocity
from observed motion and interactions. More recent benchmarks extend this
evaluation to general-purpose VLMs. PhysBench~\cite{chow2025physbench}
covers physical properties, relations, and dynamics, whereas
QuantiPhy~\cite{puyin2026quantiphy} focuses on video-based numerical
estimation with explicit textual physical priors. This setting aligns directly with ATS, which studies answer-level selective
trust for numerical predictions and uses controlled prior interventions to
probe prediction reliability. Unlike benchmark-level accuracy evaluation, our
focus is whether an estimate produced by a fixed VLM should be retained or
rejected without revising its value.

\subsection{Behavioral Reliability Estimation}

Black-box reliability methods estimate uncertainty from model behavior rather
than internal confidence. Repeated-sampling approaches use answer agreement
\cite{wang2023selfconsistency}, response inconsistency
\cite{manakul2023selfcheckgpt}, or uncertainty over semantic response clusters
\cite{kuhn2023semantic}. Although broadly applicable, these signals may assign
high confidence to consistently repeated but numerically incorrect answers.
Multimodal methods additionally measure reliability under visual or textual
perturbations \cite{zhang2024vluncertainty}, compare reference-free
hallucination signals \cite{li2024reference}, or incorporate cross-modal
interventions into uncertainty calibration \cite{padhi2025grounding}. ATS
builds on this behavioral perspective but introduces diagnostics tailored to
quantitative physical reasoning, including a counterfactual probe of
numerical-prior sensitivity.
Selective prediction converts reliability estimates into accept-or-abstain
decisions characterized by the risk--coverage trade-off
\cite{elyaniv2010foundations}. Prior work has studied post-hoc and jointly
learned rejection in classification \cite{geifman2017selective,
geifman2019selectivenet}, as well as learned or calibrated abstention in VQA
\cite{whitehead2022reliable,dancette2023improving,
eisenschlos2024selectively}. In contrast, ATS performs post-hoc selective
prediction for numerical VLM outputs using only black-box behavioral
diagnostics.

\begin{figure*}[t]
    \centering
    \includegraphics[
        width=\textwidth,
        trim=15 15 15 15,
        clip
    ]{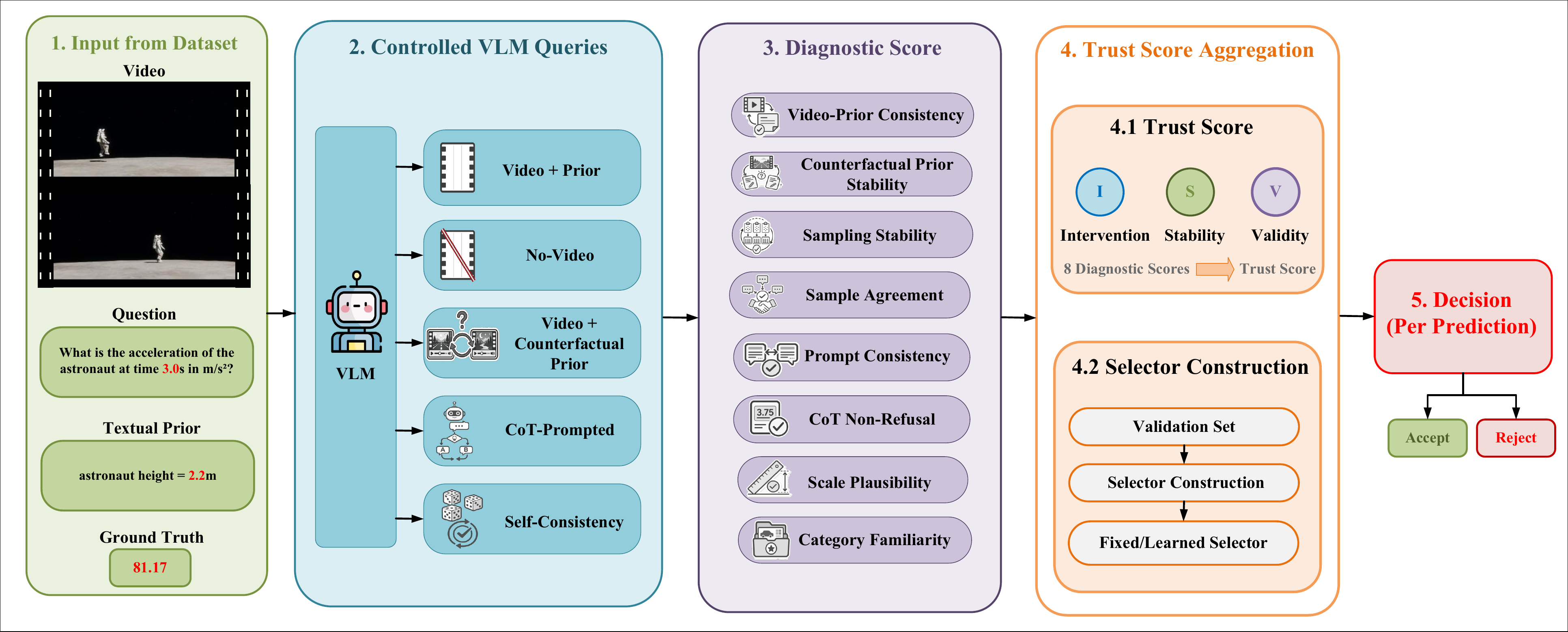}
\caption{
Overview of ATS. Controlled VLM queries produce 8 diagnostic scores
capturing intervention sensitivity, output stability, and response validity.
These scores are aggregated into a trust score, and a selector constructed on
the validation set accepts or rejects each original numerical prediction.
}
    \label{fig:stcert_workflow}
\end{figure*}

\section{Problem Formulation}
\label{sec:problem-formulation}

We study answer-level selective prediction for quantitative physical reasoning
with vision-language models (VLMs). Each example is represented as
\[
x_i=(v_i,q_i,p_i),
\]
where \(v_i\) denotes the visual observation, \(q_i\) the question, and \(p_i\)
the textual context containing a numerical prior. Given \(x_i\), a VLM \(f\)
produces an original numerical prediction
\[
\hat{y}_i=f(v_i,q_i,p_i),
\]
whose ground-truth physical quantity is denoted by \(y_i\).

Our objective is not to modify the underlying VLM or correct its original
prediction. Instead, we perform post-hoc selection to determine whether
\(\hat{y}_i\) should be accepted or rejected. Specifically, a selection method
assigns each prediction a trust score
\[
s_i=S(x_i,\hat{y}_i;f),
\]
where \(S\) may obtain additional behavioral evidence through black-box queries
to \(f\), but does not use \(y_i\) at inference time. Given a selection
threshold \(\tau\), the corresponding selector is
\[
g_\tau(x_i,\hat{y}_i)
=
\mathbb{I}[s_i\geq\tau],
\]
where \(g_\tau=1\) indicates acceptance and \(g_\tau=0\) indicates rejection.
Varying \(\tau\) yields different trade-offs between the fraction of retained
answers and their predictive quality.

We measure numerical prediction quality using Mean Relative Accuracy
(MRA)~\cite{puyin2026quantiphy}.
Let
\[
\mathcal{C}=\{0.1,0.2,\ldots,0.9,0.95\},
\]
and let \(\epsilon_y>0\) be a small constant for denominator stabilization. The
per-example MRA of prediction \(\hat{y}_i\) is defined as
\[
\MRA(\hat{y}_i,y_i)
=
\frac{1}{|\mathcal{C}|}
\sum_{\theta\in\mathcal{C}}
\mathbb{I}
\left[
\frac{|\hat{y}_i-y_i|}
{\max\{|y_i|,\epsilon_y\}}
<
1-\theta
\right].
\]
This metric averages prediction correctness across multiple relative-error
tolerances, assigning higher values to estimates that are closer to the ground
truth.

For a test set of \(N\) examples, the coverage of \(g_\tau\) is the fraction of
accepted predictions:
\[
\Cov(\tau)
=
\frac{1}{N}
\sum_{i=1}^{N}
g_\tau(x_i,\hat{y}_i).
\]
For any threshold with nonzero coverage, the retained-answer MRA is
\[
\MRAret(\tau)
=
\frac{
\sum_{i=1}^{N}
\MRA(\hat{y}_i,y_i)\,
g_\tau(x_i,\hat{y}_i)
}{
\sum_{i=1}^{N}
g_\tau(x_i,\hat{y}_i)
}.
\]
The corresponding selective risk is
\[
R(\tau)
=
1-\MRAret(\tau).
\]
A reliable selector should achieve lower selective risk at a given coverage,
or equivalently higher coverage at a given risk level.

\section{Methods}
\subsection{Overview}

As illustrated in Figure~\ref{fig:stcert_workflow}, ATS is a post-hoc framework for answer-level
selective trust in numerical physical reasoning. For each example, the VLM
is evaluated under five controlled conditions: Video + Prior, Prior Only,
Video + Counterfactual Prior, CoT-Prompted, and Self-Consistency Sampling.
The resulting responses are converted into eight behavioral diagnostic
scores and aggregated into an overall trust score. The validation set is then used to
construct an answer-level selector that determines whether the original
numerical prediction should be accepted for downstream use.


\subsection{Controlled VLM Probing}
\label{sec:controlled-probing}

ATS constructs answer-level diagnostic scores by probing the VLM under five controlled query settings. For each example
$x=(v,q,p)$, we first obtain the standard Video+Prior prediction
\[
\hat{y}_{\mathrm{full}} = f(v,q,p),
\]
which serves as the original numerical answer to be assessed.

We then obtain a prior-only prediction without visual input,
\[
\hat{y}_{\mathrm{prior}} = f(\varnothing,q,p),
\]
and a counterfactual-prior prediction,
\[
\hat{y}_{\mathrm{cf}} = f(v,q,p^{\mathrm{cf}}),
\]

where \(p^{\mathrm{cf}}\) changes only the numerical value in the textual
prior while preserving the video, question, and target physical quantity.
The intervention is designed as a diagnostic perturbation rather than a new
task instance: the original prediction remains the answer under assessment,
and the perturbed query is used only to measure whether the model is
excessively attracted to the altered textual cue.

To assess prompt sensitivity, we additionally query the VLM using a
chain-of-thought prompt,
\[
\hat{y}_{\mathrm{cot}}
=
f(v,q,p;\pi_{\mathrm{cot}}),
\]
where $\pi_{\mathrm{cot}}$ denotes the chain-of-thought prompting instruction.

Finally, to assess sampling stability, we draw $M$ stochastic
generations from the standard Video+Prior input,
\[
\hat{y}_{\mathrm{sc}}^{(m)}
\sim f(v,q,p;\xi_m),
\qquad m=1,\ldots,M,
\]
where $\xi_m$ denotes the sampling randomness. These repeated outputs
are summarized by their median,
\[
\tilde{y}_{\mathrm{sc}}
=
\operatorname{median}_{m=1,\ldots,M}
\hat{y}_{\mathrm{sc}}^{(m)},
\]
and their dispersion, which is later used to compute the
sampling stability score.

\subsection{Diagnostic Score Construction}
\label{sec:diagnostic-score}
A single numerical prediction is insufficient to determine whether an answer
should be trusted. ATS therefore probes the same VLM through a set of controlled black-box
queries. By comparing the model outputs across these
conditions, ATS constructs eight complementary diagnostic scores. Each score describes a specific behavioral phenomenon rather than a
calibrated probability of correctness. Instead, the scores provide complementary
answer-level evidence that is subsequently aggregated into an overall trust
score.

The \textbf{video-prior consistency score} compares the standard
video-conditioned prediction with the prior-only prediction in log-magnitude
space, where
\(\ell(z)=\log\!\left(\max\{|z|,\epsilon_{\log}\}\right)\) and
\(\epsilon_{\log}>0\):
\[
\Svp=
\frac{1}{
1+\left|
\ell(\hat{y}_{\mathrm{full}})
-\ell(\hat{y}_{\mathrm{prior}})
\right|
}.
\]
A larger \(\Svp\) indicates closer agreement in log magnitude between the
Video + Prior and Prior-Only predictions.

The \textbf{counterfactual prior stability score} tests whether the prediction remains
close to the original video-conditioned answer when the textual prior is
perturbed. Let
\[
\rho
=
\frac{\mathrm{num}(p^{\mathrm{cf}})}
{\mathrm{num}(p)}
\]
denote the numerical scale ratio between the counterfactual and original
priors. We compare the counterfactual prediction against both the original
video-conditioned prediction and an idealized prior-tracking reference:
\[
d_{\mathrm{ref}}
=
\left|
\ell(\hat{y}_{\mathrm{cf}})
-
\ell(\hat{y}_{\mathrm{full}})
\right|,
\]
\[
d_{\mathrm{track}}
=
\left|
\ell(\hat{y}_{\mathrm{cf}})
-
\ell(\rho\hat{y}_{\mathrm{full}})
\right|.
\]
The counterfactual prior stability score is
\[
\Scps
=
\frac{
d_{\mathrm{track}}
}{
\max\!\left\{
\zeta,\,
d_{\mathrm{track}}+d_{\mathrm{ref}}
\right\}
},
\]
where \(\zeta>0\) is a denominator floor. A larger \(\Scps\) indicates
that the counterfactual prediction remains closer to the original
video-conditioned prediction than to the idealized prior-tracking reference.

For self-consistency sampling, let
\(\{\hat{y}^{(m)}\}_{m=1}^{M}\) denote the repeated numerical generations,
and let
\[
\mathcal{V}
=
\left\{
m\in\{1,\ldots,M\}
:
\hat{y}^{(m)}\in\mathbb{R}_{>0}
\right\}
\]
denote the set of valid positive predictions.

The \textbf{sampling stability score} measures the log-space dispersion
of these valid repeated predictions:
\[
\sigma_{\log}
=
\operatorname{Std}
\!\left(
\left\{
\ell\!\left(\hat{y}^{(m)}\right)
\right\}_{m\in\mathcal{V}}
\right),
\qquad
s_{\mathrm{rep}}
=
\frac{1}{1+\sigma_{\log}}.
\]
A larger \(s_{\mathrm{rep}}\) indicates that the valid repeated predictions
are more tightly concentrated in log space.

The \textbf{sample agreement score} measures the proportion of repeated
generations that produce a valid positive numerical prediction:
\[
a_{\mathrm{valid}}
=
\frac{1}{M}
\sum_{m=1}^{M}
\mathbb{I}\!\left[
\hat{y}^{(m)}\in\mathbb{R}_{>0}
\right],
\qquad
s_{\mathrm{agr}}
=
a_{\mathrm{valid}}.
\]
A larger \(s_{\mathrm{agr}}\) indicates that a greater proportion of repeated
generations produce valid positive numerical predictions. Thus,
\(s_{\mathrm{agr}}\) represents a valid-output rate rather than agreement
around a dominant numerical value.

The \textbf{prompt consistency score} measures the numerical
consistency across the Prior-Only, Video + Prior, and CoT-Prompted
conditions. Let \(\hat{y}_{\mathrm{cot}}\) denote the numerical prediction
extracted from the CoT-Prompted response. We define
\[
\begin{aligned}
\sigma_{\mathrm{prompt}}
&=
\operatorname{Std}\!\left(
\ell(\hat{y}_{\mathrm{prior}}),
\ell(\hat{y}_{\mathrm{full}}),
\ell(\hat{y}_{\mathrm{cot}})
\right),\\
s_{\mathrm{prompt}}
&=
\frac{1}{1+\sigma_{\mathrm{prompt}}}.
\end{aligned}
\]
A larger \(s_{\mathrm{prompt}}\) indicates greater numerical consistency
across these query conditions.

The \textbf{scale-plausibility score} measures whether the original
video-conditioned prediction has an extreme log-magnitude:
\[
s_{\mathrm{scale}}
=
\exp\!\left(
-\frac{
\left|\ell(\hat{y}_{\mathrm{full}})\right|
}{
\kappa_{\mathrm{scale}}
}
\right),
\]
where \(\kappa_{\mathrm{scale}}>0\) is a fixed scale constant. We set
\(\kappa_{\mathrm{scale}}=5\) in all experiments.
A larger \(s_{\mathrm{scale}}\) indicates that the prediction has a less
extreme log-magnitude under this scale heuristic.

The \textbf{category familiarity score} measures how frequently the event
family of an example occurs in a fixed reference pool. Let \(h_i\) denote
the event family of example \(i\), let \(c_{h_i}\) denote its empirical
count in the reference pool, and let \(N\) denote the pool size. We first
define the frequency-based event-family novelty risk as
\begin{equation}
r_{i,\mathrm{unk}}
=
1-
\frac{\log(1+c_{h_i})}{\log(1+N)},
\label{eq:r_unk}
\end{equation}
and convert it into positive familiarity evidence:
\[
s_{i,\mathrm{fam}}
=
1-r_{i,\mathrm{unk}}.
\]
A larger \(s_{i,\mathrm{fam}}\) indicates that the corresponding event family
occurs more frequently in the fixed reference pool.

The \textbf{CoT non-refusal score} measures whether the CoT-Prompted response
avoids a predefined set of refusal patterns:
\[
s_{\mathrm{valid}}
=
\mathbb{I}\!\left[
R_{\mathrm{cot}}
\notin
\mathcal{R}_{\mathrm{refusal}}
\right],
\]
where \(R_{\mathrm{cot}}\) is the CoT-Prompted response and
\(\mathcal{R}_{\mathrm{refusal}}\) is the predefined set of refusal patterns.
A larger \(s_{\mathrm{valid}}\) indicates that the CoT-Prompted response does
not match a predefined refusal pattern. Specifically,
\(s_{\mathrm{valid}}=1\) denotes non-refusal and
\(s_{\mathrm{valid}}=0\) denotes refusal.


\subsection{Trust-Score Aggregation}
\label{sec:trust-aggregation}

The diagnostic scores capture complementary behavioral evidence about each
prediction. ATS organizes the eight scores into three groups:
intervention-related, stability-related, and validity-related scores:
\begin{equation}
\begin{aligned}
\mathbf{s}_{\mathrm{int},i}
&=
\left[
s_{\mathrm{vp},i},
s_{\mathrm{cps},i}
\right]^{\top},\\
\mathbf{s}_{\mathrm{stab},i}
&=
\left[
s_{\mathrm{rep},i},
s_{\mathrm{agr},i},
s_{\mathrm{prompt},i}
\right]^{\top},\\
\mathbf{s}_{\mathrm{val},i}
&=
\left[
s_{\mathrm{fam},i},
s_{\mathrm{scale},i},
s_{\mathrm{valid},i}
\right]^{\top}.
\end{aligned}
\label{eq:signal_groups}
\end{equation}
The intervention-related scores characterize how predictions respond to
controlled changes in the textual prior. The stability-related scores measure
consistency across stochastic generations and prompting conditions. The
validity-related scores provide complementary evidence concerning event
familiarity, numerical scale, and response validity. After clipping each score
to the interval $[0,1]$, ATS computes the aggregate trust score as
\begin{equation}
\tau_i
=
\mathbf{w}_{\mathrm{int}}^{\top}
\mathbf{s}_{\mathrm{int},i}
+
\mathbf{w}_{\mathrm{stab}}^{\top}
\mathbf{s}_{\mathrm{stab},i}
+
\mathbf{w}_{\mathrm{val}}^{\top}
\mathbf{s}_{\mathrm{val},i}.
\label{eq:trust_score}
\end{equation}

We evaluate three aggregation strategies. {Prior-weighted ATS}, our
default setting, uses a set of manually specified weights. These weights are
specified heuristically and fixed before test evaluation.
{Equal-weight ATS} assigns the same weight to all eight scores, providing
a control without manually specified priorities. {Learned ATS} uses a
lightweight logistic-regression model to learn the score combination from the
validation set. Finally, we conduct a weight-sensitivity analysis by perturbing the prior-weighted
weights and recomputing retained-answer MRA at fixed coverage levels and AURC.
This analysis examines whether ATS remains stable under moderate changes to
the aggregation weights. Complete analysis are provided in the
\textit{supplementary material.}

\subsection{Selector Construction}
\label{sec:certificate-calibration}

ATS converts the diagnostic trust score into an answer-level accept/reject
rule using the validation set. The main implementation uses event-conditional
selection rules. Let \(\mathcal{H}\) denote the set of event families, and let
\(h_i=H(x_i)\in\mathcal{H}\) denote the event family assigned to example \(i\),
as determined by its physical inference type and video type. For each event
family \(h\) and candidate trust threshold \(\lambda\in\Lambda_h\), define the
candidate rule parameters
\[
\gamma_{h,\lambda}
=
(
\alpha_{h,\lambda},
\beta_{h,\lambda},
\eta_{h,\lambda},
\lambda
).
\]
The validation procedure selects \(\widehat{\lambda}_h\in\Lambda_h\), the corresponding
family-specific rule is
\[
\widehat{\gamma}_h
=
(
\widehat{\alpha}_h,
\widehat{\beta}_h,
\widehat{\eta}_h,
\widehat{\lambda}_h
).
\]
The deployed selector for example \(i\) is
\begin{equation}
\begin{aligned}
g_{\widehat{\gamma}_{h_i}}(x_i,\hat{y}_i)
=
\mathbb{I}\Big[
& s_{i,\mathrm{vp}}\ge \widehat{\alpha}_{h_i}
\;\wedge\;
s_{i,\mathrm{cps}}\ge \widehat{\beta}_{h_i} \\
&\wedge\;
r_{i,\mathrm{unk}}\le \widehat{\eta}_{h_i}
\;\wedge\;
\tau_i\ge \widehat{\lambda}_{h_i}
\Big].
\end{aligned}
\label{eq:certificate_selector}
\end{equation}

 In the practical empirical
procedure, candidate operating points are evaluated on the validation split
using empirical coverage and selective risk, where selective risk is defined as
one minus the mean per-example \(\MRA\) among accepted examples. Let
\(\Lambda_h\) denote the finite set of candidate operating points for event
family \(h\), and let
\[
K=\sum_{h\in\mathcal H}|\Lambda_h|
\]
denote the total number of candidates. We additionally compute the
Hoeffding-style upper-confidence screening statistic
\[
\overline R_h(\gamma_{h,\lambda})
=
\widehat R_h(\gamma_{h,\lambda})
+
\sqrt{
\frac{\log(K/\delta)}
{2n_{h,\lambda}}
},
\]
where \(n_{h,\lambda}>0\) denotes the number of accepted validation examples
and \(\delta\in(0,1)\) is the family-wise failure probability. In practice,
candidate rules are evaluated on the validation set using empirical coverage,
empirical selective risk, and the Hoeffding-style screening statistic, which
penalizes rules supported by few accepted examples. The selected rule is then
fixed before test evaluation, where its coverage and selective risk are
reported as empirical performance measures. The theoretical development,
proof, and additional analyses are provided in the
\textit{supplementary material}.


\subsection{Test-Time Selection}
\label{sec:test-time-certification}

During validation procedure, ATS uses the validation set to construct one selector for
each event family. Each selector contains family-specific acceptance
thresholds and is fixed before test evaluation. For each test example, ATS
applies the selector corresponding to its event family to the original
numerical prediction. The prediction is accepted only if all prespecified selection conditions are
satisfied. Otherwise, it is rejected. No test labels are used, and ATS neither
revises nor replaces the original prediction.

\subsection{Theoretical Motivations}

ATS is primarily an empirical answer-selection framework. This subsection provides two analytical observations that motivate its design: a fixed-rule concentration result explaining the sample-size penalty used during validation, and a monotonicity property explaining the construction of the counterfactual-prior stability score. Because the practical ATS rules are constructed and evaluated using the same validation split, these observations are not presented as finite-sample guarantees for the experimental procedure.

We first characterize the selective risk of a finite collection of
prespecified event-conditional accept/reject rules. 
Let
\(
L(\hat{y},y)=1-\MRA(\hat{y},y)\in[0,1]
\)
denote the bounded numerical loss. For each event family \(h\), let
\(\Gamma_h=\{\gamma_{h,1},\ldots,\gamma_{h,K_h}\}\) be a finite family of
complete accept/reject rules. Each rule specifies all diagnostic-score
definitions, aggregation weights, thresholds, event-family structure, and
acceptance conditions. For \(\gamma\in\Gamma_h\), define the conditional
selective risk as
\[
R_h(\gamma)
=
\mathbb{E}\!\left[
L(\hat{Y},Y)
\mid
g_{\gamma}(X,\hat{Y})=1,\ H(X)=h
\right].
\]
Let \(\widehat R_h(\gamma)\) denote the corresponding empirical selective risk
computed over \(n_{h,\gamma}>0\) accepted examples in the risk-calibration
sample.
\begin{proposition}[Conditional finite-class selective-risk bound]
Assume that calibration units are conditionally i.i.d. within each event family
and that the finite family of complete candidate rules
\(\Gamma=\bigcup_{h\in\mathcal H}\Gamma_h\) is fixed independently of the
risk-calibration sample. Then, with probability at least \(1-\delta\),
simultaneously for every candidate with \(n_{h,\gamma}>0\),
\[
R_h(\gamma)
\le
\widehat R_h(\gamma)
+
\sqrt{\frac{\log(K/\delta)}{2n_{h,\gamma}}}.
\]
Hence, any data-dependent selection from this prespecified family inherits its
corresponding simultaneous upper bound.
\end{proposition}

If each \(\lambda\in\Lambda_h\) indexes a prespecified complete rule, then
\(K=\sum_{h\in\mathcal H}|\Lambda_h|\), matching the finite-class correction
used in the practical screening statistic.
The proposition controls average loss among accepted examples, not
answer-specific correctness or a prediction interval. In the practical
two-way procedure, some acceptance thresholds are derived adaptively from the
same validation data used to evaluate empirical selective risk. The resulting
rule is fixed before test evaluation, but the proposition's independence
condition is not satisfied. We therefore interpret the Hoeffding-style
quantity \cite{hoeffding1963probability} as a conservative screening statistic and the test results as
empirical selective-trust evidence rather than a formal finite-sample
guarantee. Further analysis is provided in the \textit{supplementary material.}
\begin{proposition}[Monotonicity of counterfactual prior stability]
Assume \(\rho>0\), \(\rho\neq1\), and that all relevant magnitudes exceed the
log floor \(\epsilon_{\log}\). Consider
\(\hat{y}_{\mathrm{cf}}=\rho^\alpha\hat{y}_{\mathrm{full}}\),
\(\alpha\in[0,1]\), where \(\alpha=0\) corresponds to reference stability and
\(\alpha=1\) corresponds to full prior tracking. Let \(A=\log\rho\neq0\). Then
\[
s_{\mathrm{cps}}(\alpha)
=
\frac{(1-\alpha)|A|}{\max\{\zeta,|A|\}}.
\]
Thus \(s_{\mathrm{cps}}(\alpha)\) decreases strictly with the degree of prior
tracking, with
\(s_{\mathrm{cps}}(0)=|A|/\max\{\zeta,|A|\}\) and
\(s_{\mathrm{cps}}(1)=0\).
\end{proposition}

\section{Experiments}
\label{sec:experiments}

\subsection{Dataset and task}

We evaluate ATS on the official validation split of QuantiPhy
\cite{puyin2026quantiphy}. 
Following the original benchmark, we measure numerical answer quality using
Mean Relative Accuracy (MRA) \cite{puyin2026quantiphy}. Beyond evaluating
numerical estimation itself, we consider an answer-level selective-prediction
setting in which ATS determines whether each VLM prediction should be accepted
or rejected. To the best of our knowledge, no other publicly available video
benchmark directly combines real-world physical-quantity estimation with the
annotations required for this evaluation. We therefore focus our experiments
on QuantiPhy.


\subsection{Baselines}
We compare Prior-weighted ATS with SC-only, three Generic UQ baselines,
Equal-weight ATS, Learned ATS, and Oracle ranking. SC-only ranks
predictions by agreement across repeated generations
\cite{wang2023selfconsistency}. Generic UQ uses only
\(s_{\mathrm{rep}}\), \(s_{\mathrm{agr}}\), and \(s_{\mathrm{prompt}}\):
the mean and rank-fusion variants aggregate these scores directly, whereas
Learned Generic UQ fits a standardized logistic-regression aggregator on the
validation set using a binary pseudo-correctness target indicating whether the
per-example MRA is at least \(0.4\). Equal-weight ATS assigns uniform weights
to the full diagnostic set. Learned ATS fits the same aggregator using all
diagnostic scores. Oracle ranking orders predictions by ground-truth answer
quality and serves as an unattainable upper bound.

\subsection{Experiment Results}
\label{sec:main-results}
This section presents the main results on ATS's selective performance. The \textit{supplementary material} provides further evaluation protocols, cross-backbone and failure-mode analyses, diagnostic studies, ablations, robustness checks, statistical tests, calibration results, and computational-cost analyses.

\begin{figure}[t]
    \centering
    \includegraphics[width=\columnwidth]
    {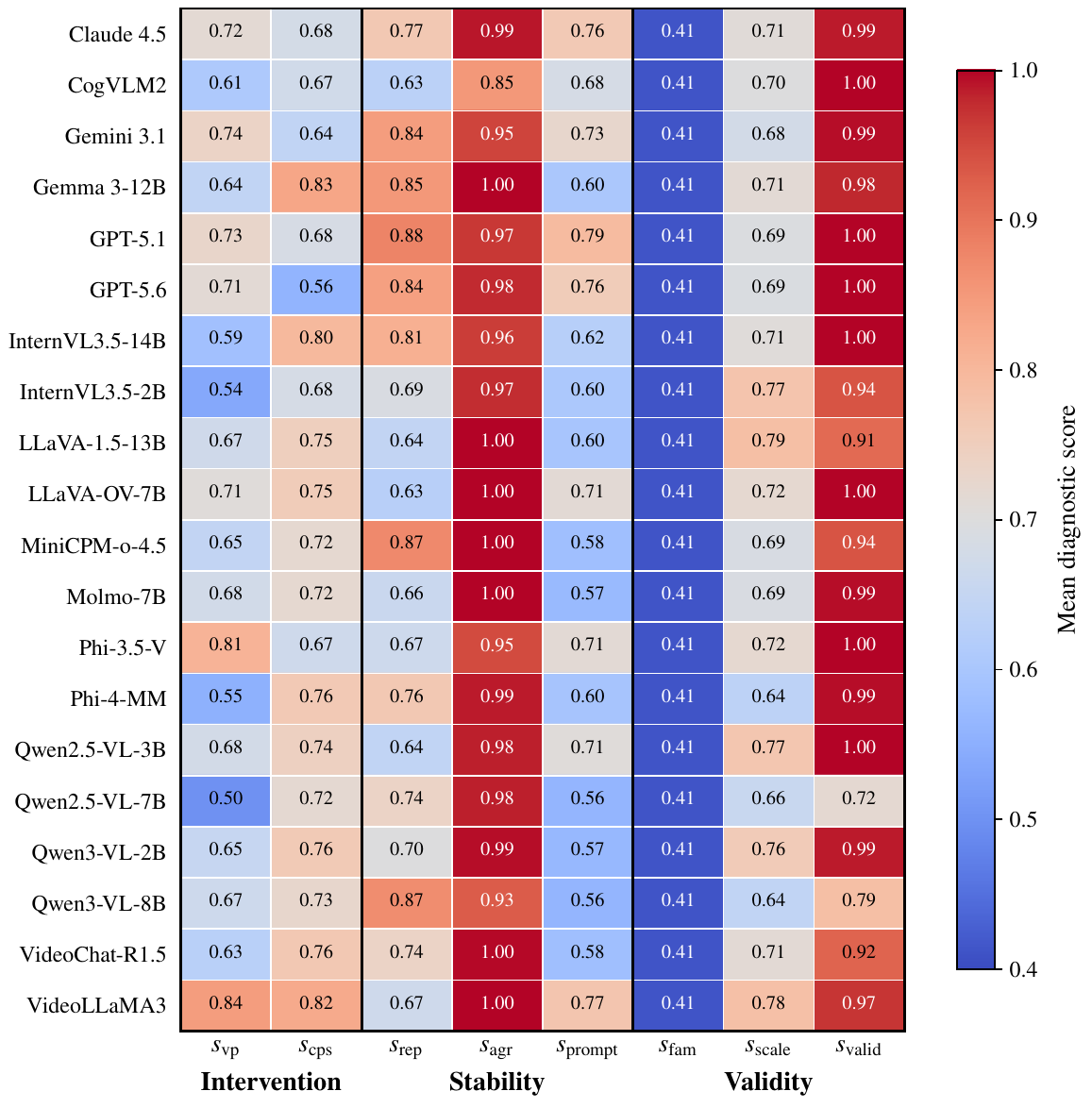}
\caption{
Cross-backbone diagnostic profiles across 20 VLMs. Each cell reports the
mean value of an answer-level diagnostic score. The figure shows how
intervention-, stability-, and validity-related scores vary across
backbones.
}
    \label{fig:cross_backbone_scores}
\end{figure}

\begin{table}[t]
\centering
\footnotesize 
\setlength{\tabcolsep}{1.5pt} 
\renewcommand{\arraystretch}{1.02}

\begin{tabular*}{\columnwidth}{@{\extracolsep{\fill}}lcccc@{}}
\toprule
& \multicolumn{2}{c}{Stable-but-wrong}
& \multicolumn{2}{c}{Prior tracking} \\
\cmidrule(lr){2-3}
\cmidrule(lr){4-5}
Selector
& \shortstack{Reject\\$\uparrow$}
& \shortstack{Correct\\retain $\uparrow$}
& \shortstack{Wrong\\reject $\uparrow$}
& \shortstack{Correct\\retain $\uparrow$} \\
\midrule

SC-only
& 0.013
& \textbf{0.986}
& 0.654
& 0.491 \\

Generic UQ
& 0.271
& 0.830
& 0.638
& \textbf{0.523} \\

Equal-weight ATS
& \textbf{0.424}
& 0.723
& 0.906
& 0.248 \\

Prior-weighted ATS
& \underline{0.409}
& 0.727
& \textbf{0.958}
& 0.111 \\

Learned ATS
& 0.379
& 0.777
& 0.574
& 0.512 \\

\bottomrule
\end{tabular*}
\caption{
Failure-case rejection and correct-case retention at 50\% coverage,
averaged across 20 backbones. 
}
\label{tab:failure_modes_main}

\end{table}

\begin{figure}[t]
    \centering
    \includegraphics[width=0.9\columnwidth]
    {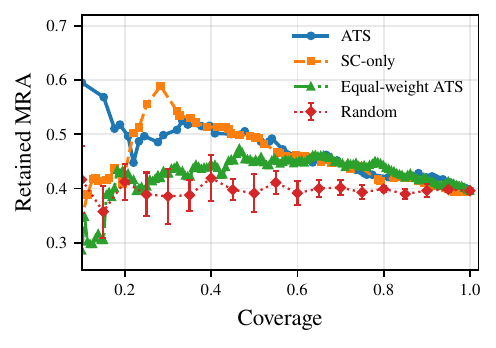}
    \caption{
    Coverage-retained-MRA curves on Qwen2.5-VL-7B. 
    }
    \label{fig:selective_curve}
\end{figure}

\subsection{Cross-backbone Diagnostic Profiles}
Figure~\ref{fig:cross_backbone_scores} summarizes the 8 ATS scores
across 20 VLM backbones. Repeated-output stability and validity-related
scores are generally high, whereas intervention- and prompt-related scores,
particularly \(s_{\mathrm{vp}}\), \(s_{\mathrm{cps}}\), and
\(s_{\mathrm{prompt}}\), show greater cross-backbone variation. We use these
profiles as descriptive behavioral evidence to motivate the targeted
failure-mode analysis, rather than as a direct performance comparison. More detailed analyses are provided in the \textit{supplementary material}.

\subsection{Cross-backbone Selective Comparison}
Across the 20 backbones, Prior-weighted ATS has mean paired effects of
$+0.0192$, $+0.0023$, and $-0.0037$ in MRA at 30\%, 50\%, and 70\%
coverage, respectively, relative to SC-only. For AURC, where a positive
difference favours ATS, the mean effect is $+0.0133$, with positive effects
on 14 of 20 backbones. We therefore treat the aggregate comparison as a
descriptive cross-backbone consistency check rather than evidence of universal
superiority. AURC shows the most consistent positive direction, whereas
fixed-coverage effects remain small and backbone dependent. More analyses of the cross-backbone experiments are provided in the \textit{supplementary material.}

\subsection{Targeted Failure-mode Analysis}
Overall selective-ranking metrics do not reveal whether a selector identifies predictions that remain consistent but numerically wrong.
Table~\ref{tab:failure_modes_main} evaluates rejection of stable-but-wrong
and prior-tracking cases at 50\% coverage, together with retention of matched
correct cases. Equal-weight and Prior-weighted ATS reject \(42.4\%\) and
\(40.9\%\) of stable-but-wrong cases, compared with \(27.1\%\) for Generic
UQ and \(1.3\%\) for SC-only. Prior-weighted ATS also rejects \(95.8\%\) of wrong
prior-tracking cases. In the table, bold indicates the best deployable result in each column,
while underlining marks the second-highest stable-but-wrong rejection rate. These rejection gains come with lower
correct-case retention and must therefore be interpreted as
rejection-retention trade-offs rather than unqualified improvements.
Nevertheless, they show that controlled interventions reveal answer-level behavioral evidence that self-consistency alone may not capture.

\subsection{Representative Qwen2.5-VL-7B Results}
Figure~\ref{fig:selective_curve} and Table~\ref{tab:baselines} summarize the
Qwen2.5-VL-7B results. The curves show how ATS and the main baseline selectors
perform across different coverage level. SC-only performs best at 30\% and 50\% coverage, while
Generic UQ rank fusion achieves the lowest AURC. Prior-weighted ATS remains
competitive across operating points and changes only modestly under random
weight perturbations. Removing \(s_{\mathrm{vp}}\) reduces MRA@50 from 0.4889
to 0.4734 and increases AURC from 0.5300 to 0.5406, suggesting that the
\(s_{\mathrm{vp}}\) contributes to ATS performance on this backbone. Together with Table~\ref{tab:failure_modes_main}, these results show that ATS
provides intervention-based evidence beyond that captured by Generic UQ,
particularly to detect stable-but-wrong and prior-tracking failures.

\begin{table}[t]
\centering
\small
\setlength{\tabcolsep}{3pt}

\begin{tabular}{lcccc}
\toprule
Method & @30\% & @50\% & @70\% & AURC $\downarrow$ \\
\midrule
\multicolumn{5}{l}{\textit{Trust-score baselines}} \\

SC-only
& \textbf{0.5429}
& \textbf{0.4989}
& 0.4426
& 0.5338 \\

Equal-weight ATS
& 0.4374
& 0.4441
& 0.4444
& 0.5655 \\

Prior-weighted ATS
& 0.5187
& 0.4889
& 0.4476
& 0.5300 \\

Learned ATS
& 0.4069
& 0.4120
& 0.4028
& 0.5977 \\

Prior-weighted ATS w/o $s_{vp}$
& 0.4879
& 0.4734
& 0.4352
& 0.5406 \\

Prior-weighted ATS w/o $s_{cps}$
& 0.4780
& 0.4886
& \textbf{0.4671}
& 0.5308 \\

\midrule
\multicolumn{5}{l}{\textit{Generic uncertainty baselines}} \\

Generic UQ mean
& 0.4677
& 0.4526
& 0.4533
& 0.5151 \\

Generic UQ rank fusion
& 0.4909
& 0.4709
& 0.4504
& \textbf{0.5125} \\

Learned Generic UQ
& 0.5063
& 0.4605
& 0.4387
& 0.5593 \\

\midrule
\multicolumn{5}{l}{\textit{Weight sensitivity}} \\

Random perturb. mean
& 0.5192
& 0.4881
& 0.4492
& 0.5306 \\

Random perturb. 10th pct.
& 0.5109
& 0.4804
& 0.4437
& 0.5363 \\

\bottomrule
\end{tabular}
\caption{
Comparison of selective-ranking baselines, ATS variants, score
ablations, and weight perturbations on Qwen2.5-VL-7B.
}
\label{tab:baselines}
\end{table}

\section{Conclusion}

We presented ATS, a post-hoc framework for answer-level
selective prediction in quantitative physical reasoning with
VLMs. ATS integrates complementary behavioral evidence
from controlled interventions, repeated-generation stability,
and response validity, without requiring access to model
internals or additional fine-tuning. Across 20 VLM backbones,
SC-only and Generic UQ remain strong baselines on aggregate
selective-ranking metrics. Targeted analyses show that ATS
rejects substantially more stable-but-wrong and prior-tracking
cases, two failure modes that self-consistency sampling or
generic uncertainty estimation alone may not fully capture.
These gains come with reduced retention of matched correct
predictions and therefore represent a rejection--retention
trade-off rather than a uniform improvement. Overall, our
results position intervention-based diagnostics as a
complementary source of answer-level uncertainty evidence
and motivate future selectors that better leverage such
diagnostic scores to improve aggregate selective performance.


\bibliography{aaai2027}


\end{document}